\documentclass[11pt,a4paper]{article}
\usepackage[hyperref]{emnlp2020}
\usepackage{mathptmx} 
\usepackage{latexsym}

\usepackage{paralist}

\usepackage{microtype}

\aclfinalcopy 

\title{COVCOR20 at WNUT-2020 Task 2: An Attempt to Combine Deep Learning and Expert rules}

\author{Ali H{\"u}rriyeto\u{g}lu \\
  Ko\c{c} University \\
  \texttt{ahurriyetoglu@ku.edu.tr} \\\\
 \bf Osman Mutlu \\
  Ko\c{c} University \\ 
  \texttt{omutlu@ku.edu.tr} \And 
  Ali Safaya \\ \newline
  Ko\c{c} University \\
  \texttt{asafaya@ku.edu.tr} \\\\
  \bf Erdem Y\"{o}r{\"u}k \\
  Ko\c{c} University \\ 
  \texttt{eryoruk@ku.edu.tr} \\ \And
  \bf Nelleke Oostdijk \\ 
  Radboud University  \\ 
  \texttt{n.oostdijk@let.ru.nl} \\ }

\date{}

\begin{document}
\maketitle
\begin{abstract}
In the scope of WNUT-2020 Task 2, we developed various text classification systems, using deep learning models and one using linguistically informed rules. While both of the deep learning systems outperformed the system using the linguistically informed rules, we found that through the integration of (the output of) the three systems a better performance could be achieved than the standalone performance of each approach in a cross-validation setting. However, on the test data the performance of the integration was slightly lower than our best performing deep learning model. These results hardly indicate any progress in line of integrating machine learning and expert rules driven systems. We expect that the release of the annotation manuals and gold labels of the test data after this workshop will shed light on these perplexing results.
\end{abstract}


\section{Introduction}

The COVID-19 pandemic urged various science disciplines to do their best so as to contribute to understanding and relieving its impact. Thus, scholars and practitioners working on information sciences have been dedicating significant effort to help. Collecting and analyzing data published on social media platforms have become the focus in this respect. We joined the community that aims at organizing data collected from social media (in this use case: Twitter), as informative and uninformative. The WNUT-2020 Task 2 considers tweets about recovered, suspected, confirmed and death cases as well as location or travel history of the cases as informative. All other tweets are considered to be uninformative. The organizers did not share an annotation manual nor was a baseline system made available, presumably to prevent use of any other manually annotated data and to encourage broad participation respectively~\cite{Nguyen+20a}.\footnote{\url{http://noisy-text.github.io/2020/covid19tweet-task.html}, accessed on September 4, 2020.}

The effort was managed in terms of a shared task, in which the organizers share a dataset that consists of annotated tweets and conduct the evaluation of the submissions. The task requires the participating teams to develop short-text classification systems that facilitate the training and development data to generalize to the test set they release. Although the gold labels of the training and development data were available to the participants, neither the gold labels of the test data nor the annotation guidelines for any part of the data were shared with the participants. Moreover, the test instances were unknown to the participating teams. They were hidden in a larger dataset. Each team was allowed to submit only two outputs of the systems they developed for classifying tweets on the Codalab page of the task.\footnote{\url{https://competitions.codalab.org/competitions/25845}, accessed on September 4, 2020.} The highest score in terms of F1 positive class of each team was used to rank them in the leaderboard.

Integrating automatically created machine learning based (ML) models with manually formulated rules to tackle a text classification task promises the best of both worlds. We pursued this goal by integrating the output of two deep learning models and a rule-based system under the team name COVCOR20. Although the integration slightly improves the total performance on the training and development sets in a cross-validation setting, the overall performance on the test data turned out to be slightly worse than our best ML system. Our best submission was ranked 22nd among 55 teams. The integration of our systems would be ranked 27th if its score were used as the final score for our team.

The deep learning models and the rule-based system are introduced in Sections~\ref{sec:deeplearn} and~\ref{sec:rulebased} respectively. Next, the Section~\ref{sec:integrate} describes how we integrate the output of these systems. Then Section~\ref{sec:results} provide the results and their discussion. Finally, we conclude this report and share our future plans continuing in this line of research in Section~\ref{sec:conclude}.

\section{Deep learning models}
\label{sec:deeplearn}
We created various deep learning models by merging the training and development data released by the organizers. We applied k-fold cross-validation by splitting the data into five parts. The average performance in terms of F1 of five optimization and test iterations is used to compare these models. The hyper-parameter optimization is focused on F1 of the positive class. Table~\ref{modelperfs} provides performances of the models we tested. CNNText, BiLSTM, and BERT-CNN were borrowed from~\citet{Safaya+20}, BERT is the standard model from~\citet{Devlin+19}, and BERTweet is the model created by~\citet{Nguyen+20b}. 

We developed a novel model that we  call Fused-1 and Fused-2 that has a modified version of CharRNN\footnote{\url{https://github.com/alisafaya/char-rnn.pytorch}, acccessed on September 6, 2020.} is a character language model that consists of two LSTM layers and is pretrained on 112 thousand tweets related to COVID19 which we collected. In Fused-1 CharRNN is applied in parallel with BERT, while in Fused-2 it is used in parallel with BERTweet. Just as in BERT-CNN, we feed the output of the last hidden layer of CharRNN to a Convolutional Neural Network (CNN). We also have another CNN that takes BERT or BERTweet output as its input. Upon this, we concatenate the output of these two CNNs (the first one using BERT or BERTweet as embedder and the latter using CharRNN as embedder), and we feed it to a fully connected layer to get the final output\footnote{All details of the models and the  optimization can be found on the repository \url{https://github.com/emerging-welfare/covid19-tweet}, which we have created for this shared task}.

\begin{table}
\centering
\begin{tabular}{lcc}
\hline \textbf{Model} & \textbf{F1} & \textbf{Std} \\ \hline
CNNText & .8543 & .0116 \\
BiLSTM & .8595 & .0089 \\
BERT & .9432 & .0066 \\
BERT-CNN & .9478 & .0060 \\
BERTweet & .9499  & .0029 \\
Fused-1 & .9552 & .0022 \\
Fused-2 & .9560 & .0024 \\
\hline
\end{tabular}
\caption{\label{modelperfs} Model performances in terms of F1 and Standard Deviation (Std).}
\end{table}

We used Fused-2 as it is the best performing model in the aforementioned cross-validation setting. Moreover, we observed that the average number of the sentences in wrongly predicted instances is larger than the correctly predicted cases using the best performing deep learning model. Therefore, we generated a second output for each tweet by predicting each sentence separately. If at least one sentence in a tweet is predicted as informative, the prediction for the whole tweet becomes positive, i.e. informative. The precision decreased, the recall increased, and the F1 slightly decreased in this setting.

\section{Rule-based system}
\label{sec:rulebased}

The rule-based system uses a set of handcrafted rules constructed by an expert for the task at hand and a task-specific lexicon. The system is designed to identify only instances of the positive class, which for this particular task are the tweets that are considered to be informative. The rules describe, in terms of linguistic patterns, the salient part(s) of these tweets. Whenever a rule fires, the tweet will be labelled as informative; tweets for which no rule is found to apply are considered non-informative by default. A similar rule-based approach has been used successfully in other shared tasks (e.g. \citet{Hurriyetoglu+17}; \citet{Oostdijk+17}; \citet{Oostdijk+19}). There, the approach has been shown to yield consistently high precision across different datasets, while recall generally falls short of that yielded by ML approaches.

In this case the lexicon we compiled has 779 entries. Entries here are of the form: word type – semantic/syntactic word class – (optionally) label, e.g. 

\begin{tabular}{lc}
  case N informative & death N informative \\
  first NUMord & covid19 Ncorona \\
  new ADJ & confirmed ADJ
\end{tabular}

In order to have maximum control over the strings matched, in most cases entries are delimited by word boundary markers (\textbackslash b), indicating the beginning and the end of a word or multi-word expression. An exception was made with the entries referring to the virus (including for example, covid19, covid-19, coronavirus) where not including the initial word boundary marker allows for the matching of instances where the word is part of a hashtag.

The rules set comprises 409 rules. They describe how the lexical entries can combine to form larger strings, the length of which in the present case varies between 2 to 7 words. Example rules are\footnote{The asterisk indicates the element from which the string inherits its label.} \\

\begin{tabular}{l}
    NUMord *N  \\
    ADJ  Ncorona *N \\
    NUMord  ADJ  *N \\
\end{tabular} 

which account for instances like \textit{first case}, \textit{first confirmed death}, and \textit{new covid19 deaths}. One of the strengths of the rule-based approach is that the rules generalize beyond the instances observed in the training set, provided that the individual words appear in the lexicon.

The performance achieved with the rule-based system was as follows: on the training set precision was .845, recall .893 and F1 86.9; on the validation set precision was .77, recall .852 and F1 .809. 
We are able increase the precision at the expense of decreasing recall by requiring the number of detected spans to be equal or more than a certain threshold. For instance, the precision and recall become .91 and .53 on average if we set this threshold to be 2.


\section{Integrating outputs}
\label{sec:integrate}
We have shown that ML and rule-based approaches yield complementary output for text classification tasks in various settings in our previous work~\cite{Hurriyetoglu+17,Oostdijk+17,Oostdijk+19}. Therefore, we continued to work along this line by focusing on integrating the outputs of these systems in the scope of this shared task. However, voting and applying OR and AND operations on the predictions did not outperform the best deep learning model Fused-2. The only integration that perform slightly better was giving precedence to the rule-based system in case the two output versions of Fused-2 conflict and the rule-based system predicts a tweet as uninformative. In case the rule-based system predicts it informative, we used prediction of the best system, which is Fused-2 on all text of a tweet. This adjustment increased F1 from .9560 to .9581 in our cross-validation setting.

\section{Results}
\label{sec:results}

We submitted the best performing deep learning model, which is Fused-2 using all text of a tweet and the best performing integration. The deep learning model yielded .8887 and the integration yielded 0.8856 F1. Our team ranked 22 with these scores.\footnote{All results of the competition are on: \url{https://competitions.codalab.org/competitions/25845\#results}} The best overall F1 score in the shared task was .9096.

The decrease of the scores from around .9560 on development set to .8887 on the test set shows that rule-based system could have had a chance by itself. However, the restriction of the submission count to be 2 did not allow us to evaluate this.


\section{Conclusion and future work}
\label{sec:conclude}

We have presented our effort in the scope of a shared task that aims at pushing the state of the art for classifying short-texts (i.e. tweets in the reported use case), as informative or uninformative. 

We could extend the training set with cluster mining using Relevancer~\cite{Hurriyetoglu+16}, use the rule-based system to extend the training set~\cite{Hurriyetouglu19c}, or use the rule-based system to generate fine-grained data that can be used in a multi-task setting.

\section*{Acknowledgements}
The authors from Koç University are funded by the European Research Council (ERC) Starting Grant 714868 awarded to Dr. Erdem Yörük for his project Emerging Welfare.

\bibliographystyle{acl_natbib}
\bibliography{emnlp2020}

\end{document}